\documentclass[11pt,a4paper]{article}
\usepackage[hyperref]{naaclhlt2019}
\usepackage{times}
\usepackage{latexsym}
\usepackage{graphicx}
\usepackage{subcaption}
\usepackage{url}
\usepackage{helvet}  %Required
\usepackage{courier}  %Required
\usepackage{amsmath}
\usepackage{xcolor }
\usepackage{amssymb}
\usepackage{float}
\usepackage{caption}
\usepackage{subcaption}
\usepackage{algorithm}
\usepackage{algorithm}
\usepackage{booktabs}  
\usepackage[noend]{algpseudocode}
\usepackage[bottom]{footmisc}
\aclfinalcopy % Uncomment this line for the final submission
\def\aclpaperid{843} %  Enter the acl Paper ID here

\title{Latent Code and Text-based Generative Adversarial Networks for Soft-text Generation}

\author{Md. Akmal Haidar, Mehdi Rezagholizadeh, Alan Do-Omri and Ahmad Rashid   \\
Huawei Noah's Ark Lab, Montreal Research Center, Canada \\
  {\tt md.akmal.haidar@huawei.com, mehdi.rezagholizadeh@huawei.com}  \\
  {\tt alan.do.omri@huawei.com, ahmad.rashid@huawei.com}
 \\}
  
\date{}

\begin{document}
\maketitle
\begin{abstract}
Text  generation  with  generative  adversarial networks  (GANs)  can  be  divided  into  the text-based  and  code-based  categories  according to the type of signals used for discrimination. In this work, we introduce a novel text-based approach called Soft-GAN to effectively exploit GAN setup for text generation. We demonstrate how autoencoders (AEs) can be used for  providing a continuous representation of sentences, which we will refer to as soft-text. This soft representation will be used in GAN discrimination to synthesize similar soft-texts.
We also propose hybrid latent code and text-based GAN (LATEXT-GAN\footnote{
 Accepted in 2019 Annual conference of the North American Chapter of the Association for Computational Linguistics (NAACL-HLT)
}) approaches with one or more discriminators, in which a combination of the latent code and the soft-text is used for GAN discriminations.
We perform a number of subjective and objective experiments on two well-known datasets (SNLI and Image COCO) to validate our techniques. We discuss the results using several evaluation metrics and show that the proposed techniques outperform the traditional GAN-based text-generation methods.
\end{abstract}

\section{Introduction}

Text generation is an active research area and has many real-world applications, including, but not limited to, machine translation~\cite{Bahdanau2014}, AI chat bots~\cite{li2017adversarial}, image captioning~\cite{Xu2015}, question answering and
information retrieval~\cite{Wang2017}. Recurrent neural network language models (RNNLMs)~\cite{Mikolov2010} 
is the most popular approach for text generation which rely on maximum likelihood estimation (MLE) solutions such as teacher forcing~\cite{williams1989} (i.e. the model is trained to predict the next word given all the previous predicted words); however, it is  well-known in the literature that MLE is a simplistic objective for this complex NLP task~\cite{li2017adversarial}. It is reported that MLE-based methods suffer from exposure bias~\cite{Huszár2015}, which means that at training time the model is exposed to gold data only, but at test time it observes its own predictions. Hence, wrong predictions quickly accumulate and result in poor text generation quality.  

However, generative adversarial networks (GANs)~\cite{goodfellow2014generative} which are based on an adversarial loss function suffers less from the mentioned problems of the MLE solutions. 
%GANs have a generator $G_{\theta}$ to generate synthetic samples from a prior distribution to fool a discriminator network $f_w$, which compares the synthetic data distribution $G_{\theta}(x)$ with real data distribution $p_{data}(x)$. The GAN objective can be formulated as a minimax game $\displaystyle\min_{\theta}\displaystyle\max_{w} E_{x \sim p_{data}} f_w(x) + E_{x \sim G_{\theta}} (1-f_w(x))$.   
The great success of GANs in image generation framework~\cite{salimans2016improved} motivated researchers to apply its framework to NLP applications as well. 
GANs have been extensively used recently in various NLP applications such as machine translation \cite{wu2017adversarial,yang2017improving}, dialogue models~\cite{li2017adversarial}, question answering~\cite{yang2017semi}, and natural language generation~\cite{gulrajani2017improved,rajeswar2017adversarial,press2017language,kim2017adversarially,zhang2017adversarial,Cifka2018,Spinks2018,Akmal2019,Jules2019,Ahmad2019}. 
However, applying GAN in NLP is challenging due to the discrete nature of the text. 
Consequently, back-propagation would not be feasible for discrete outputs and it is not straightforward to pass the gradients through the discrete output words of the generator. 

Traditional methods for GAN-based text generation can be categorized according to the type of the signal used for discrimination into two categories: text-based and code-based techniques. 
%The schematic of the categories are described in Figure~\ref{Intro}. 
Code-based methods such as adversarial autoencoder (AAE)~\cite{makhzani2015adversarial} and adversarially regularized AE (ARAE)~\cite{kim2017adversarially} derive a latent space representation of the text using an AE and attempt to learn data manifold of that latent space \cite{kim2017adversarially} instead of modeling text directly. Text-based solutions, such as reinforcement learning (RL) based methods or approaches based on continuous approximation of discrete sampling, focus on generating text directly from the generator. 
RL-based methods treat the distribution of GAN generator as a stochastic policy and hold the discriminator responsible for providing proper reward for  the generator's actions. However, the RL-based methods often need pre-training and are computationally more expensive compared to the methods of the other two categories. 
In the continuous approximation approach for generating text with GANs, the goal is to find a continuous approximation of the discrete sampling by using the Gumbel Softmax technique~\cite{kusner2016gans} or approximating the non-differentiable argmax operator with a continuous function~\cite{zhang2017adversarial,gulrajani2017improved}. 

In this paper, we introduce Soft-GAN as a new solution for the main bottleneck of using GAN for text generation. Our solution is based on an AE to derive a soft representation of the real text (i.e. soft-text).  This soft-text is fed to the GAN discriminator instead of the conventional one-hot representation
 used in ~\cite{gulrajani2017improved}.
%, an improved Wasserstein GAN (IWGAN) was proposed where the discriminator is used for discriminating the one-hot representation with the softmax representation of the synthesized texts. We show that the proposed Soft-GAN method  outperforms the conventional IWGAN approach.
Furthermore, we propose hybrid latent code and text-based GAN (LATEXT-GAN) approaches and show that how we can improve code-based and text-based text generation techniques by considering both signals in the GAN framework.  
% \begin{figure}[!htb]
%  \includegraphics[scale=0.57]{Fig/Intro1.png}
%   \caption{Text Generation using GANs}
%  \label{Intro}
% \end{figure}
We summarize the main contributions of this paper as:
%of this paper as following : 
\begin{itemize}
\item 
%We use an AE to derive a soft representation of the real text for text-based discrimination in GAN training instead of using a conventional one-hot representation. 
We introduce a new text-based solution Soft-GAN using the above soft-text discrimination. We also demonstrate the rationale behind this approach.
\item We introduce LATEXT-GAN approaches for GAN-based text generation using both latent code and soft-text discrimination. To the best of our knowledge, this is the first time where a GAN-based text generation framework uses both code and text-based discrimination.
\item We evaluate our methods using subjective and objective evaluation metrics. We show that our proposed approaches outperform the conventional GAN-based text generation techniques that do not need pre-training.
\end{itemize}

\section{Background}

Generative adversarial networks include two separate deep networks: a generator and a discriminator. The generator $G_{\theta}$ takes in a random variable, $z$ following a distribution $P_z(z)$ and attempt to map it to the real data distribution $P_x(x)$. The output distribution of the generator is expected to converge to the real data distribution during the training. On the other hand, the discriminator $f_w$ is expected to discern real samples from generated ones by outputting zeros and ones, respectively. 
During training, the generator and discriminator generate samples and classify them, respectively by adversarially affecting the performance of each other. %The GAN objective can be formulated as a minimax game~\cite{goodfellow2014generative}:
In this regard, an adversarial loss function is employed for training~\cite{goodfellow2014generative}: 
\begin{equation}
\begin{split}
\min_{\theta} \max_w E_{x \sim P_x(x)} f_w(x)+E_{z\sim P_z(z)}(1-f_w(G_{\theta}(z)))  
\end{split}
\end{equation} 

As stated, using GANs for text generation is challenging because of the discrete nature of text. 
%To clarify the issue, Figure~\ref{simple_GAN} depicts a simplistic architecture for GAN-based text generation. 
The main bottleneck is the \textit{argmax} operator which is not differentiable and blocks the gradient flow from the discriminator to the generator.
\begin{equation}
\begin{split}
& \min_{\theta}  E_{z\sim P_z(z)}(1-f_w(\text{argmax}(G_{\theta}(z))))  
\end{split}
\end{equation} 

%This is a two-player minimax game for which a Nash-equilibrium point should be derived. Finding the solution of this game is non-trivial and there has been a great extent of literature dedicated in this regard~\cite{salimans2016improved}. 
% \begin{figure}[H]
% \centering
% \includegraphics[scale=0.40]{Fig/simple_GAN}
% \caption{Simplistic text generator with GAN}
% \label{simple_GAN}
% \end{figure}

% As stated, using GANs for text generation is challenging because of the discrete nature of text. To clarify the issue, Figure~\ref{simple_GAN} depicts a simplistic architecture for GAN-based text generation. The main bottleneck of the design is the \textit{argmax} operator which is not differentiable and blocks the gradient flow from the discriminator to the generator.

% \begin{equation}
% \begin{split}
% & \min_G  E_{z\sim P_z(z)}[\text{log}(1-D(\text{argmax}(\text{softmax}(G(z)))))]  
% \end{split}
% \end{equation} 

\section{Related Work}
\label{Related_Work}

\textbf{Text-based Solutions} Generating text using pure GANs was inspired by improved Wasserstein GAN (IWGAN) work
%which was trained by using WGAN-gradient penalty (WGAN-GP) approach 
~\cite{gulrajani2017improved}. In IWGAN, a character level language model was developed based on adversarial training of a generator and a discriminator. Their generator is a convolution neural network (CNN) generating fixed-length texts. The discriminator is another CNN receiving 3D tensors as input sentences. 
%It determines whether the tensor is coming from the generator or sampled from the real data. 
The real sentences and the generated ones are represented using one-hot and softmax representations, respectively.
%The generator produces a softmax vector over the entire vocabulary. The discriminator is responsible for distinguishing between the one-hot representations of the real text and the softmax vector of the synthesized text. 
%A disadvantage of this technique is that the discriminator is able to tell apart the one-hot input from the softmax input very easily.
A similar approach to IWGAN was proposed in~\cite{rajeswar2017adversarial} with a recurrent neural network (RNN) based generator.
In~\cite{press2017language}, RNN is trained to generate text with GAN using curriculum learning~\cite{bengio2009curriculum}.
The TextGAN~\cite{zhang2017adversarial} method was proposed to alleviate the mode-collapsing problem by matching the high-dimensional latent feature distributions of real and synthetic sentences~\cite{salimans2016improved}. Morever, several versions of the RL-based techniques using GAN have been introduced in the literature including Seq-GAN~\cite{yu2017seqgan}, RankGAN~\cite{Lin2017}, MaliGAN~\cite{Che2017}, LeakGAN~\cite{guo2017long}, and MaskGAN~\cite{fedus2018maskgan}. 
\\
\textbf{Code-based Solutions} AEs have been exploited along with GANs in different architectures for computer vision application such as AAE~\cite{makhzani2015adversarial}, ALI~\cite{dumoulin2016adversarially}, and HALI~\cite{belghazi2018hierarchical}. 
Similarly, AEs can be used with GANs for generating text. For instance, ARAE  was proposed  in~\cite{kim2017adversarially} where it employs a discrete auto-encoder to learn continuous codes based on discrete inputs with a WGAN objective to learn an implicit probabilistic model over these codes. ARAE aims at exploiting GAN’s ability to push the generator to follow a continuous code space corresponding to the encoded real text in the AE.
The generated code is then used by the decoder to generate the synthetic texts. A different version of the ARAE method was also introduced in~\cite{Spinks2018}. In~\cite{Sandeep2018}, sentence embeddings were learned by a generator to model a pre-trained sentence representation obtained by a general-purpose sentence encoder. A temperature sweeping framework was discussed in~\cite{Massimo2018} to evaluate the text generation models over whole quality-diversity spectrum and pointed out the fundamental flaws of quality-only evaluation. The Variational autoencoders (VAEs)~\cite{Kingma2013} were also applied for text generation in~\cite{Bowman2015,hu2017toward}. %In~\cite{hu2017toward}, the text generation is controlled by learning disentangled latent representations with designated semantics. 

\section{Methodology}
\begin{figure*}
    
\begin{subfigure}{0.24\textwidth}
\includegraphics[width=\linewidth]{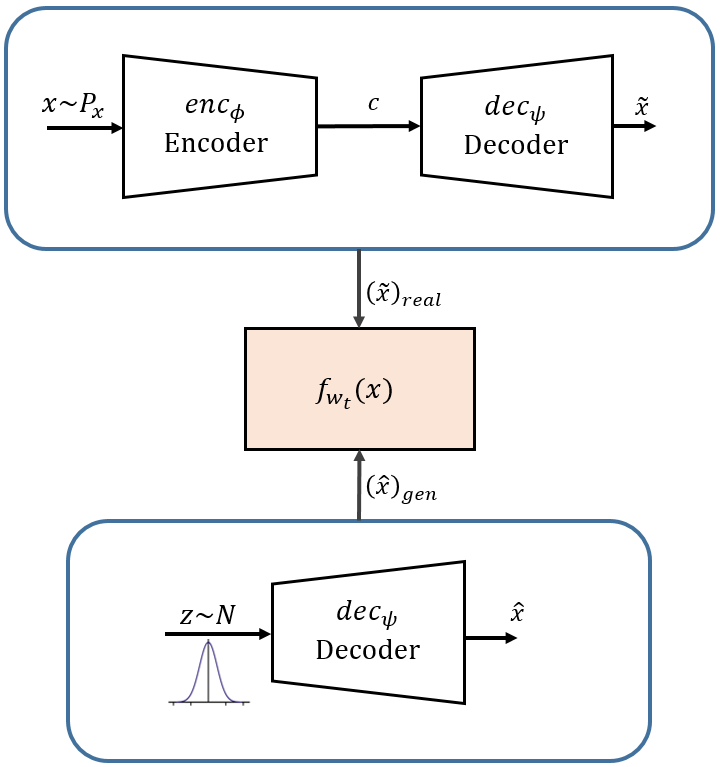}
\caption{Soft-GAN} \label{Soft-GAN}
\end{subfigure}
\hspace*{0.02cm} 
\begin{subfigure}{0.24\textwidth}
\includegraphics[width=\linewidth]{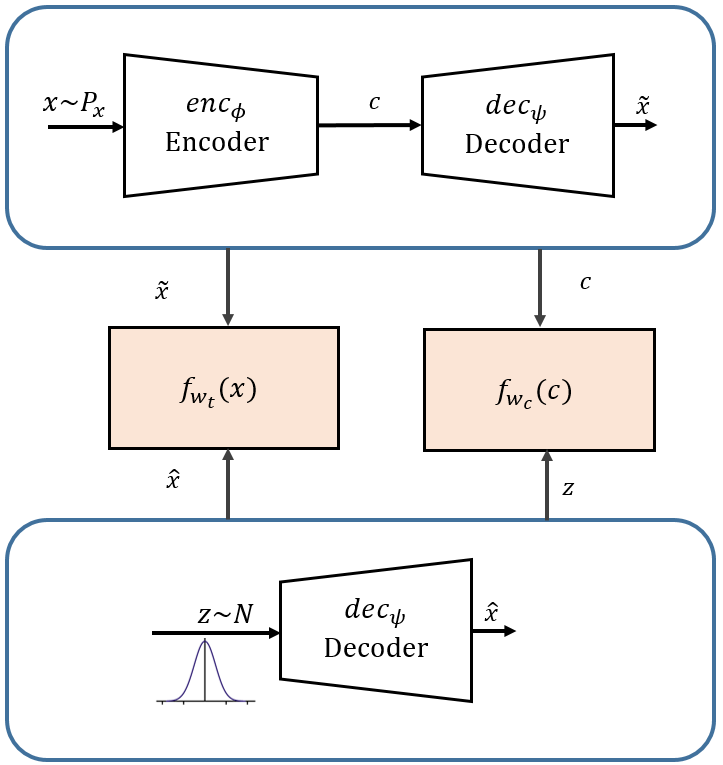}
\caption{LATEXT-GAN I} \label{AAE-Soft-text-critic}
\end{subfigure}
\hspace*{0.02cm} 
\begin{subfigure}{0.24\textwidth}
\includegraphics[width=\linewidth]{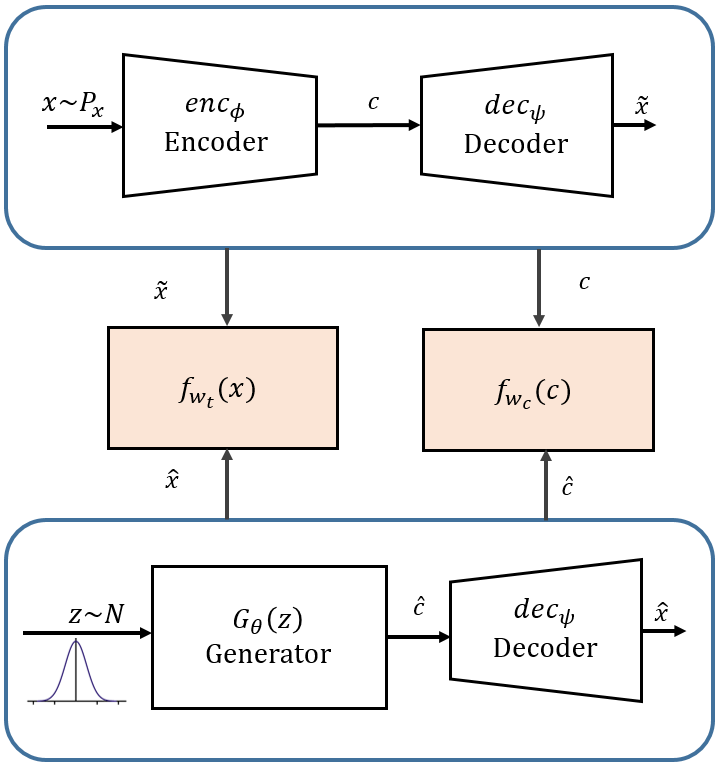}
\caption{LATEXT-GAN II} \label{ARAE-Soft-text-critic}
\end{subfigure}
\hspace*{0.02cm} 
\begin{subfigure}{0.24\textwidth}
\includegraphics[width=\linewidth]{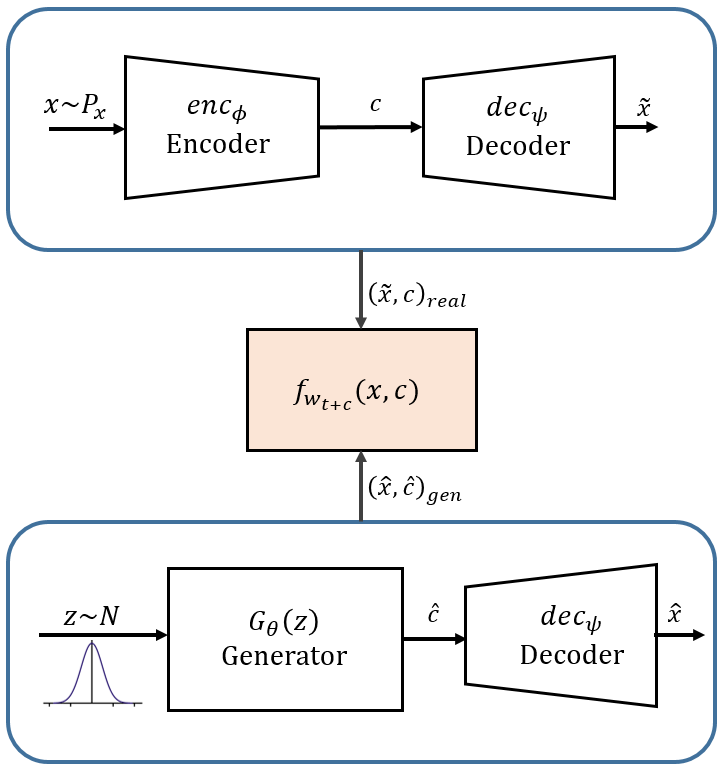}
\caption{LATEXT-GAN III} \label{ARAE-ALI}
\end{subfigure}
\caption{Proposed Approaches} \label{methods}
\end{figure*}

In this section, we introduce a new text-based solution by discriminating the reconstructed output of an AE (i.e., soft-text) with the synthesized generated text and we call it as Soft-GAN. Here,  we use the decoder of the AE as the generator to generate the synthesized texts from random noise data $z$. The model is described in Figure~\ref{Soft-GAN}. We demonstrate the rationale behind this soft-GAN approach, which is to make the discrimination task of the discriminator between the real and synthetic texts more difficult and consequently providing a richer signal to the generator. We also introduce three LATEXT-GAN approaches where both code and text-based signals will be used in the GAN framework. We introduce LATEXT-GAN I approach on top of the AAE method. LATEXT-GAN II and III approaches will be proposed based on the ARAE approach. In the LATEXT-GAN I and II techniques, we use separate discriminators for the code and text discriminations. In the LATEXT-GAN III approach, the concatenation of the synthetic code and the synthetic text tries to mimic the concatenation of the latent code and the soft-text using a discriminator. The schematic diagram of the LATEXT-GAN I, II, and III approaches are described in Figures~\ref{AAE-Soft-text-critic},~\ref{ARAE-Soft-text-critic}, and~\ref{ARAE-ALI} respectively. We share the parameters of the decoder of the AE to generate the synthesized text $\hat{x}$.
%from random data in Figures~\ref{Soft-GAN} and~\ref{AAE-Soft-text-critic}, or from the synthetic code in Figures~\ref{ARAE-Soft-text-critic} and ~\ref{ARAE-ALI}.
\begin{table}[!htb]
    \centering
  \scalebox{0.85}{
  \begin{tabular}{|l|l|}
    \hline
     $x$  & One-hot representation of the training text  \\ 
$\tilde{x}$ & Soft-text: Reconstructed output of the AE \\
$\hat{x}$  &Synthesized generated text\\ 
$\bar{x}$  & 
$[\bar{x}\sim P_{\bar{x}}] \leftarrow \alpha ~ [\tilde{x}\sim P_{\tilde{x}}] + (1-\alpha)~ [\hat{x}\sim P_{\hat{x}}]   $
\\
%&Random data samples obtained by sampling \\
%&uniformly along a line connecting pairs of \\ 
%& synthetic and real text data samples \\ 
$z \sim N$ & Random data drawn from a normal distribution\\
$c$ &Latent code representation of the training text  \\ 
$\hat{c}$ & Synthesized generated code\\ 
$\bar{c}$  &$[\bar{c}\sim P_{\bar{c}}] \leftarrow \alpha ~ [c\sim P_{c}] + (1-\alpha)~ [\hat{c}\sim P_{\hat{c}}]    $\\
$\bar{cz}$  &$[\bar{cz}\sim P_{\bar{cz}}] \leftarrow \alpha ~ [c\sim P_{c}] + (1-\alpha)~ [z\sim P_{z}]    $\\
%&Random latent code obtained by sampling \\
%& uniformly along a line connecting pairs of the \\
%& synthetic code and the encoder output\\ 
$w_{t+c}$  &Parameters of the combination of text \\
& and code-based discriminator\\ 
$w_t$  &Parameters of the text-based discriminator\\ 
$w_c$  &Parameters of the code-based discriminator\\ 
$\phi$  &Parameters of the encoder\\ 
$\psi$  &Parameters of the decoder\\ 
$\theta$  &Parameters of the generator \\ 
$\lambda$ & Gradient penalty co-efficient\\
$\nabla$ & describes gradient\\

%$\lambda_t$, $\lambda_c$, $\lambda_{t+c}$  & Gradient penalty co-efficients\\
\hline
  \end{tabular}}
  \caption{Notations that are used in this paper}
  \label{notation}
\end{table}
\vspace{-.2cm}
In order to train these approaches, we train the auto-encoder and the GAN  alternatingly by minimizing their loss functions using 
%. We train the GAN by using 
the WGAN-gradient penalty (WGAN-GP) approach (Gulrajani et al., 2017). 
In each iteration, the first step is the AE training in all of these  techniques followed by the GAN loss functions. The autoencoder can be trained by using a cross-entropy or mean-squared loss functions. The input $x$ to the AE is mapped to a latent code representation $c$ which is decoded to soft-text $\tilde{x}$. In our experiments, we train the auto-encoder using mean-squared loss $\displaystyle\min_{(\phi, \psi)} ||x-\tilde{x}||^2$.
We describe the training details of the Soft-GAN, LATEXT-GAN I, II, and III methods in the following subsections where the term critic and discriminator used interchangeably. The notations that are used in this paper are described in Table~\ref{notation}.

\subsection{Soft-GAN}
\label{soft-gansec}
As stated, in conventional text-based discrimination approach  IWGAN~\cite{gulrajani2017improved}, the real and the synthesized generated text are described by the one-hot and the softmax represenation respectively. A disadvantage of this technique is that the discriminator is able to tell apart the one-hot input from the softmax input very easily. One way to avoid this issue is to derive a continuous representation of words rather than their one-hot and train the discriminator to differentiate between the continuous representations. We use a conventional AE to replace the one-hot representation with softmax reconstructed output ($\tilde{x}$), which we refer to as soft-text. This soft-text representation is used as the real input to the discriminator. The synthetic generated text $\hat{x}$ is obtained by inputting the random noise data $z$ to the shared decoder. We define the proposed method as Soft-GAN. In each iteration, the model is trained using the following steps after the AE training step:
\begin{itemize}
\item Train the text-based discriminator $f_{w_t}$ for $k$ times and the decoder once using the  loss $L_{{critic}_t}$ to maximize the ability of the $f_{w_t}$ to discriminate between $\tilde{x}$ and $\hat{x}$: 
\begin{equation}
\begin{split}
&  L_{{\textit{critic}_t}} = \min_{(w_t,\psi)}(-E_{\tilde{x} \sim P_{\tilde{x}}} [f_{w_t} (\tilde{x})]+E_{\hat{x}\sim P_{\hat{x}} } \\  
& [f_{w_t} (\hat{x})] + \lambda E_{\bar{x}\sim P_{\bar{x}}} [(||\nabla_{\bar{x}}   f_{w_t} (\bar{x})||_2-1)^2])\\
\end{split}
\label{eq3}
\end{equation}
\vspace{-.7cm}
\item Train the decoder based on the loss $L_{gen}$ to fool the discriminator with improving the representation $\hat{x}$:
\begin{equation}
\begin{split}
& L_{\textit{gen}} =\min_{\psi}(-E_{\hat{x} \sim P_{\hat{x}}} [f_{w_t} (\hat{x})] + E_{\tilde{x} \sim P_{\tilde{x}}} \\
& [f_{w_t} (\tilde{x})])
\end{split}
\label{loss_prop1}
\end{equation}
\end{itemize}
\subsubsection{Rationale: Why Soft-GAN should Work Better than IWGAN?}
\label{Rationale}
\begin{figure}[t]
\fbox{
\begin{minipage}[t]{.445\linewidth}
\centering
\includegraphics[scale=0.5]{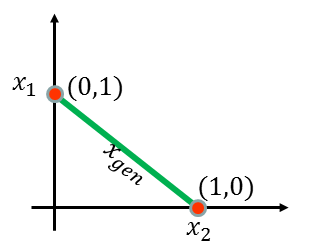}
\end{minipage}}
\fbox{
\begin{minipage}[t]{.445\linewidth}
\includegraphics[scale=0.48]{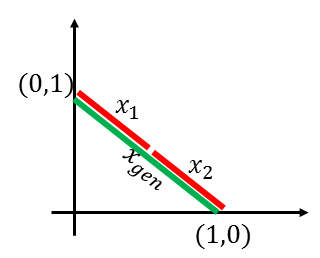}
\end{minipage}}
\caption{Locus of  the input vectors to the discriminator$ f_{w_t}$ for a two-word language; Left panel: IWGAN, Right panel: Soft-GAN }
\label{Fig_ex}
\end{figure}

Suppose we have a language of vocabulary size of two words: $x_1$ and $x_2$. In the IWGAN approach, the one-hot representation of these two words (as two points in the Cartesian coordinates) and the span of the generated softmax outputs (as a line segment connecting them) is depicted in the left panel of Figure~\ref{Fig_ex}.
As evident graphically, the task of the critic is to discriminate the points from the line connecting them, which is a rather simple very easy task, which makes it more prone to vanishing gradient. 

On the other hand, the output locus of the softGAN decoder would be two red line segments as depicted in Figure~\ref{Fig_ex} (Right panel) instead of two points (in the one-hot case). The two line segments lie on the output locus of the generator, which will make the generator more successful in fooling the critic.

\subsection{LATEXT-GAN I}
In the LATEXT-GAN I approach (Figure~\ref{AAE-Soft-text-critic}), we deploy two critics: one for the soft-text discrimination and the other for the latent code discrimination. The text-based discriminator $f_{w_t}$ is used to discriminate the soft-text output $\tilde{x}$ with the synthesized text $\hat{x}$ which is obtained by inputting the random noise data to the shared decoder. The code-based discriminator $f_{w_c}$ is used to discriminate the random noise data $z$ with the latent code $c$ in the AAE setting which was explored for image generation. In the AAE setting~\cite{makhzani2015adversarial}, the encoder enhances its representation to a prior distribution $z$.  It can be seen that the LATEXT-GAN I can be obtained by adding the above code-based discriminator $f_{w_c}$ into the Soft-GAN. In each iteration, the model is trained using the following steps after the AE training step and the  Equation~\ref{eq3} step of the Soft-GAN in section~\ref{soft-gansec}:
\begin{itemize}
\item Train the code-based discriminator $f_{w_c}$ for $k$ times using the loss $L_{{critic}_c}$ to maximize the ability of the $f_{w_c}$ to discriminate between $c$ and $z$: 
\begin{equation}
\begin{split}
&  L_{{\textit{critic}_c}} = \min_{w_c} (E_{c \sim P_{c}} [f_{w_c} (c)]-E_{z\sim P_{z} }  \\ 
&[f_{w_c} (z)]+ \lambda E_{\bar{cz}\sim P_{\bar{cz}}} [(||\nabla_{\bar{cz}}   f_{w_c} (\bar{cz})||_2-1)^2])\\
\end{split}
\end{equation}
\vspace{-.7cm}
\item Train the encoder and the decoder neural networks once with the loss $L_{gen}$ to fool the discriminators $f_{w_c}$ and $f_{w_t}$  with improving the representations  $c$ and $\hat{x}$ respectively:
\begin{equation}
\begin{split}
& L_{\text{gen}} = \min_{(\phi,\psi)} (- E_{\hat{x} \sim P_{\hat{x}}} [f_{w_t} (\hat{x})] + E_{\tilde{x}\sim P_{\tilde{x}}}  \\
& [f_{w_t} (\tilde{x})] + E_{z \sim P_{z}} [f_{w_c} (z)] - E_{c \sim P_{c}} [f_{w_c} (c)])  
\end{split}
\end{equation} 
\end{itemize}
\subsection{LATEXT-GAN II}
The LATEXT-GAN II approach (Figure~\ref{ARAE-Soft-text-critic}) is similar to the LATEXT-GAN I approach except the training of the code-based discriminator $f_{w_c}$ is done as the ARAE training. 
The critic $f_{w_c}$ is used to discriminate the synthetic code $\hat{c}$ with the latent code $c$. Here, the synthetic code is formed by using the ARAE method~\cite{Spinks2018}. For each iteration in the model training, the AE training step and the Equation~\ref{eq3} step of the Soft-GAN in section~\ref{soft-gansec} are carried out first. Then the following two steps are performed:
\begin{itemize}
\item Train the code-based discriminator $f_{w_c}$ for $k$ times and the encoder once using the loss $L_{{critic}_c}$ to maximize the ability of the $f_{w_c}$ to discriminate between $c$ and $\hat{c}$: 
\begin{equation}
\begin{split}
&  L_{{\textit{critic}_c}} = \min_{(w_c,\phi)} (- E_{c \sim P_{c}} [f_{w_c} (c)]+E_{\hat{c}\sim P_{\hat{c}} }  \\ 
&[f_{w_c} (\hat{c})]+ \lambda E_{\bar{c}\sim P_{\bar{c}}} [(||\nabla_{\bar{c}}   f_{w_c} (\bar{c})||_2-1)^2])\\
\end{split}
\end{equation}
\vspace{-.5cm}
\item Train the generator and the decoder neural networks once using the loss $L_{gen}$ to fool the discriminators $f_{w_t}$  and $f_{w_c}$ with improving the representations  $\hat{x}$ and $\hat{c}$ respectively:
\begin{equation}
\begin{split}
& L_{\text{gen}} = \min_{(\theta,\psi)} (- E_{\hat{x} \sim P_{\hat{x}}} [f_{w_t} (\hat{x})] + E_{\tilde{x}\sim P_{\tilde{x}}}  \\
& [f_{w_t} (\tilde{x})] - E_{\hat{c} \sim P_{\hat{c}}} [f_{w_c} (\hat{c})] + E_{c \sim P_{c}} [f_{w_c} (c)])  
\end{split}
\end{equation}     
\end{itemize}
\subsection{LATEXT-GAN III}
In the third approach (Figure~\ref{ARAE-ALI}), the combination of latent code $c$ generated by ARAE~\cite{Spinks2018} and the soft-text output $\tilde{x}$ of an AE is used to signal the discriminator. We performed this combination by getting inspiration from an Adversarially Learned Inference (ALI) paper (Dumoulin et al., 2016) introduced for image generation. We call it as LATEXT-GAN III. Here, the discriminator $f_{w_{t+c}}$ tries to determine which combination of the samples derive from the latent code and the soft-text, ($\tilde{x}$,$c$), and which ($\hat{x},\hat{c}$) are generated from the noise $z$. After the AE training step  $\displaystyle\min_{(\phi, \psi)} ||x-\tilde{x}||^2$, the LATEXT-GAN III model is trained using the next two steps in each iteration:
\begin{itemize}
\item Train the discriminator $f_{w_{t+c}}$ for $k$ times, the encoder and the decoder once using the loss $L_{{critic}_{t+c}}$ to maximize the ability of the discriminator network to discriminate between ($\tilde{x}$,$c$) and ($\hat{x},\hat{c}$): 
\begin{equation}
\begin{split}
&  L_{{\textit{critic}_{t+c}}} = \min_{(w_{t+c},\phi,\psi)} (- E_{(\tilde{x},c) \sim P_{\tilde{x}},P_{c}} [f_{w_{t+c}} (\tilde{x},c)] \\
& + E_{(\hat{x},\hat{c}) \sim P_{\hat{x}},P_{\hat{c}} } 
[f_{w_{t+c}} (\hat{x},\hat{c})] + \\
&\lambda E_{(\bar{x},\bar{c}) \sim P_{\bar{x}},P_{\bar{c}}} [(||\nabla_{(\bar{x},\bar{c})}   f_{w_{t+c}} (\bar{x},\bar{c})||_2-1)^2])\\
\end{split}
\end{equation}   
\vspace{-.7cm}
\item Train the generator and the decoder once based on $L_{gen}$ to fool the discriminator $f_{w_{t+c}}$ with improving the representation ($\hat{x},\hat{c}$):
\begin{equation}
\begin{split}
& L_{\text{gen}} = \min_{(\theta,\psi)} (- E_{(\hat{x},\hat{c}) \sim P_{\hat{x}},P_{\hat{c}}} [f_{w_{t+c}} (\hat{x},\hat{c})]\\
& +E_{(\tilde{x},c)\sim P_{\tilde{x}},P_{c} } [f_{w_{t+c}} (\tilde{x},c)])  
\end{split}
\end{equation}     
\end{itemize}

\vspace{-0.7cm}
\section{Experiments}
\label{results}
\subsection{Dataset and Experimental Procedures}
We do our experiments on two different datasets: 
the Stanford Natural Language Inference (SNLI) corpus~\footnote{https://github.com/aboev/arae-tf/tree/master/data\_snli}, which contains 714,667 sentences for training and 13323 sentences for testing, and the Image COCO~\footnote{http://cocodataset.org} dataset's image caption annotations, where we sample~\footnote{https://github.com/geek-ai/Texygen/tree/master/data} 10,000 sentences as training set 
and another 10,000 as test set~\cite{zhu2018texygen}. We perform  word-based experiments. For the SNLI dataset, we use a vocabulary size of 10000 words and use the maximum sentence length of size 15. For the COCO dataset, we use a vocabulary size of 5000 and perform experiments using the maximum sentence length of sizes 15 and 20.
We train a simple AE using one layer with 512 LSTM cells~\cite{Hochreiter1997} for both the encoder and the decoder. For decoding, the output from the previous time step is used as the input to the next time step. We use the hidden code $c$ from the last time step of the encoder and applied as an additional input at each time step of decoding. We normalize the code and then added an exponentially decaying noise before decoding. The greedy search approach is applied to get the best output. We train the auto-encoder using Adam~\cite{Diederik2014} optimizer with learning rate = 0.001, $\beta_1$= 0.9, and $\beta_2$= 0.999. We use CNN-based generator and discriminator with residual blocks~\cite{gulrajani2017improved}. The $tanh$ function is applied on the output of the ARAE generator~\cite{kim2017adversarially}. We train the generator and the discriminator using Adam optimizer with learning rate = 0.0001, $\beta_1$= 0.5, and $\beta_2$= 0.9. We do not apply any kind of attention mechanisms~\cite{Bahdanau2014,Zhang2018} and pre-training~\cite{zhu2018texygen} in our experiments.
We use the WGAN-GP~\cite{gulrajani2017improved} approach with 5 discriminator updates for every generator update and a gradient penalty co-efficient of $\lambda$=10 unlike a setup in~\cite{zhu2018texygen}. For the AAE-based experiments, we normalize the data drawn from a prior distribution $z$. We train the models for 200000 iterations where in each iteration we sample a random batch and train the networks of the models. 
\vspace{-0.2cm}
\subsection{Quantitative Evaluation}
We use the frequently used BLEU metric ~\cite{papineni2002} 
to evaluate the word similarity between sentences and the perplexity to evaluate our techniques. We calculate BLEU-n scores for n-grams without a brevity penalty~\cite{zhu2018texygen}.
The results with the best BLEU-n scores in the synthesized generated texts are reported. To calculate the BLEU-n scores, we generate ten batches of sentences as candidate texts, i.e. 640 sentences and use the entire test set as reference texts. As the GAN-based models usually suffer from mode collapse (i.e., generating same samples over and over again), evaluating models by only BLEU metric is not appropriate. So, we also calculate recently proposed self-BLEU scores for the COCO dataset using maximum sentence length of size 20 and 10k synthetic sentences to evaluate the diversity~\cite{zhu2018texygen}. Using one synthetic  sentence as hypothesis and others as reference, the BLEU is calculated for every synthetic sentence, and define the average BLEU score as the self-BLEU~\cite{zhu2018texygen}.
%For each generated  sentence, we compute the BLEU using the sentence as hypothesis and the rest of the generated sentences as the reference. The average BLUE score is described as the Self-BLEU score of the generated sentences.
A higher self-BLEU score describe less diversity. For the perplexity evaluations, we generate 100k and 10k sentences for the SNLI and the COCO datasets respectively using the models of the last iteration.

\subsubsection{BLEU-n Scores evaluation}
The BLEU score results for the n-grams of the synthesized texts are depicted in Table~\ref{bleu-snli} and ~\ref{bleu-coco1} with maximum sentence length of 15 for the SNLI and the COCO datasets respectively. We also report experimental results with a longer maximum sentence length of 20 using the COCO dataset to differentiate the effectiveness of code and text-based solutions (in Table~\ref{bleu-coco2}). Furthermore, we report the BLEU and self-BLEU score results of our proposed approaches in Table~\ref{bleu-coco3} and~\ref{bleu-coco4} respectively for the COCO dataset to compare with the results of the existing approaches reported in~\cite{zhu2018texygen}.  
%and to do a comparison with existing results~\cite{zhu2018texygen} to our proposed approaches in Table~\ref{bleu-coco3} and~\ref{bleu-coco4}. %Figure~\ref{BLEU-Self-BLEU_10k}. 

\begin{table}[!htb]
  \small
  \centering
  \begin{tabular}{|l|l|l|l|l|}
  \hline
    Model       & B-2&B-3 & B-4 & B-5   \\
    \hline
    
    AAE           & 0.797      & 0.614          & 0.449        &  0.294    \\
    ARAE           &0.73      &     0.575      &      0.431   &   0.297   \\
    IWGAN  &0.70& 0.518   & 0.369   & 0.246\\
\hline
    Soft-GAN  &0.849 &0.648  & 0.446 & 0.252 \\
    LATEXT-GAN I  &\textbf{0.87}& \textbf{0.679}& \textbf{0.508} & 0.336\\
    LATEXT-GAN II  &0.793& 0.631  & 0.48 & \textbf{0.338}\\
    LATEXT-GAN III  &0.782& 0.617  & 0.466 & 0.33\\
    \hline
    \end{tabular}
  \caption{BLEU-n (B-n) scores results using SNLI dataset and 640 synthetic sentences}
  \label{bleu-snli}
\end{table}
\vspace{-0.5cm}
\begin{table}[!htb]
  \small
  \centering
  \begin{tabular}{|l|l|l|l|l|}
  \hline
    Model       & B-2&B-3 & B-4 & B-5   \\
    \hline
    AAE      & 0.733 & 0.477 & 0.284 & 0.156     \\
    ARAE     & 0.676& .457  & 0.287   & 0.172     \\
    IWGAN    & 0.636  & 0.417 & 0.258 & 0.155 \\
\hline
    Soft-GAN  &\textbf{0.781} & 0.492  & 0.296 & 0.155\\
    LATEXT-GAN I  & 0.758&\textbf{0.496} & 0.294 & 0.155 \\
    LATEXT-GAN II  &0.707& 0.489  & \textbf{0.316} & \textbf{0.198} \\
    LATEXT-GAN III  &0.701 & 0.483  & 0.311 & 0.193\\
    \hline
    
  \end{tabular}
  \caption{BLEU-n (B-n) scores results for COCO dataset with maximum sentence length of size 15 and 640 synthetic sentences}
  \label{bleu-coco1}
\end{table}
\vspace{-0.3cm}
\begin{table}[!htb]
  \small
  \centering
  \begin{tabular}{|l|l|l|l|l|}
  \hline
    Model       & B-2&B-3 & B-4 & B-5   \\
    \hline
    AAE      &0.751  & 0.475 & 0.287   & 0.167     \\
    ARAE     &0.665 &  0.447 & 0.279   &   0.162   \\
    IWGAN    &0.669  & 0.454 & 0.294   & 0.178 \\

    \hline
    Soft-GAN  &\textbf{0.799} &\textbf{0.520}&\textbf{0.317}& \textbf{0.190}\\
    LATEXT-GAN I  &0.77& 0.503  & 0.314 & 0.185\\
    LATEXT-GAN II  &0.687& 0.456  & 0.283 & 0.174\\
    LATEXT-GAN III  &0.680& 0.466  & 0.292 & 0.178\\
    \hline
    \end{tabular}
  \caption{BLEU-n (B-n) scores results for COCO dataset with maximum sentence length of size 20 and over 640 synthetic sentences}
  \label{bleu-coco2}
\end{table}

% \begin{figure*}[!htb]

% \begin{minipage}[t]{.5\linewidth}
% \includegraphics[scale=1]{Fig/BLEU_10k.png}
% \end{minipage}
% \begin{minipage}[t]{.5\linewidth}
% \includegraphics[scale=1]{Fig/Self-BLEU_10k.png}
% \end{minipage}
% \caption{Results for COCO dataset using sequence length 20 and 10k generated sentences. \textbf{Left:} BLEU (B-n) scores. \textbf{Right:} Self-BLEU (SB-n) scores}
% \label{BLEU-Self-BLEU_10k}
% \end{figure*}

From tables~\ref{bleu-snli},~\ref{bleu-coco1}, and~\ref{bleu-coco2}, we can see that our proposed approaches outperform the standalone code (AAE or ARAE) and text-based (IWGAN) solutions.  
For the maximum sentence length of size 15 experiments, the LATEXT-GAN I is better than LATEXT-GAN II and III for shorter length text (e.g., 2,3-grams). The performance of the LATEXT-GAN II and III degrades with increasing maximum sentence length to 20. This is because for longer sequence length experiments, the hidden code of the last time step might not be able to keep all the information from the earlier time steps. On the other hand, the LATEXT-GAN I and the Soft-GAN improve their performance with increasing maximum sentence length to 20. This might be because of the encoder enhances its representation better to the prior distribution, $z$ from which the text is generated. Furthermore, the Soft-GAN outperforms all the proposed LATEXT GAN approaches. 

We also compare our proposed approaches with TextGAN~\cite{zhang2017adversarial}, some RL-based approaches (SeqGAN~\cite{yu2017seqgan}, RankGAN~\cite{Lin2017}, MaliGAN~\cite{Che2017}) and MLE approach described in a benchmark platform~\cite{zhu2018texygen} where they apply pre-training before applying adversarial training. We evaluate the BLEU and Self-BLEU score results on 10k synthetic sentences using the maximum sentence length of size 20 for the COCO dataset with a vocabulary of size 5000 as in~\cite{zhu2018texygen}. The BLEU and the self-BLEU score results are reported in Table~\ref{bleu-coco3} and~\ref{bleu-coco4} respectively. From Table~\ref{bleu-coco3}, 
%in the left and the right side of the Figure~\ref{BLEU-Self-BLEU_10k} respectively. 
%From Figure~\ref{BLEU-Self-BLEU_10k}, 
it can be noted that our proposed approaches show comparable results to the RL-based solutions for the BLEU score results. 
We can also see that our proposed LATEXT-GAN III approach gives lower self-BLEU scores in Table~\ref{bleu-coco4}. From the above experimental results, we can note that LATEXT-GAN III can generate real-like and more diverse sentences compare to some approaches reported in~\cite{zhu2018texygen} and our other proposed approaches.

\begin{table}[!htb]
  \small
  \centering
  %\scalebox{0.91}{
  \begin{tabular}{|l|l|l|l|l|}
  \hline
    Model       & B-2&B-3 & B-4 & B-5   \\
    
    \hline
        TextGAN &0.593  & 0.463 & 0.277   & 0.207 \\
    SeqGAN &0.745  & 0.498 & 0.294 & 0.180 \\
    RankGAN &0.743  & 0.467 & 0.264 & 0.156 \\
    MaliGAN &0.673  & 0.432 & 0.257 & 0.159 \\
    MLE &0.731  & 0.497 & 0.305 & 0.189 \\
    \hline
    Soft-GAN  &0.787 &0.496  & 0.286 & 0.150\\
    LATEXT-GAN I  &0.736& 0.447  & 0.258 & 0.146\\
    LATEXT-GAN II  &0.672& 0.430  & 0.257 & 0.147\\
    LATEXT-GAN III  &0.660& 0.435  & 0.260 & 0.149\\

    \hline
    
  \end{tabular}
  \caption{\textit{BLEU-n (B-n)} scores results for COCO dataset using sequence length 20 and 10k generated sentences}
  \label{bleu-coco3}

\end{table}

\begin{table}[!htb]

  \small
  
  \centering
  %\scalebox{0.91}{
  \begin{tabular}{|l|l|l|l|l|}
  \hline

    Model       & B-2&B-3 & B-4 & B-5   \\
    \hline
    
        TextGAN &0.942  & 0.931 & 0.804   & 0.746 \\
    SeqGAN &0.950  & 0.840 & 0.670   & 0.489 \\
    RankGAN &0.959  & 0.882 & 0.762 & 0.618 \\
    MaliGAN &0.918  & 0.781 & 0.606 & 0.437 \\
    MLE &0.916  & 0.769 & 0.583 & 0.408 \\
    \hline
    Soft-GAN  &0.988 &0.950  & 0.847 & 0.612\\
    LATEXT-GAN I  &0.991& 0.957  & 0.854 & 0.613\\
    LATEXT-GAN II  &0.896& 0.755  & 0.523 & 0.263\\
    LATEXT-GAN III  &\textbf{0.874}& \textbf{0.706} &\textbf{0.447} & \textbf{0.205}\\
    
    \hline
  \end{tabular}
  \caption{Self-BLEU scores results for COCO dataset using sequence length 20 and 10k generated sentences}
  \label{bleu-coco4}
\end{table}

\subsubsection{Perplexity Evaluation}
The forward and reverse perplexities of the LMs trained with maximum sentence length of 15 and  20 using the SNLI and the COCO datasets respectively are described in Table~\ref{ppl-snli}. 
The forward perplexities (F-PPL) are calculated by training an RNN language model~\cite{zaremba2015} on real training data and evaluated on the synthetic samples. This measure describe the fluency of the synthetic samples. We also calculate the reverse perplexities (R-PPL) by training an RNNLM on the synthetic samples and evaluated on the real test data. We can easily compare the performance of the LMs by using the forward perplexities while it is not possible by using the reverse perplexities as the models are trained using the synthetic samples with different vocabulary sizes. The perplexities of the LMs using real data are 16.01 and 67.05 for the SNLI and the COCO datasets respectively reported in F-PPL column. From the tables, we can note the models with lower forward perplexities (higher fluency) for the synthetic samples tend to have higher reverse perplexities except the AAE-based models ~\cite{Cifka2018} and/or the IWGAN. 
%This might be  because the R-PPL is evaluated using the LMs that are trained on synthetic sentences which are not sufficient. 
%and they might have ungrammatical sentences, which give the higher reverse perplexities on real test data.
%The lower reverse perplexities of the models than the forward perplexities are might be because of the sentences are less diverse or insufficient vocabularies. 
The forward perplexity for the IWGAN is the worst which means that the synthetic sentences of the IWGAN model are not fluent or real-like sentences. For the SNLI dataset, we can note that the LATEXT-GAN II and III approaches can generate more fluent sentences than the other approaches. For the COCO dataset, it can be seen that the forward perplexity of the LATEXT-GAN I (51.39) is far lower than the real data (67.05) which means the model suffers from mode-collapse. The Soft-GAN, the LATEXT-GAN II and III approaches suffer less from the mode-collapse.  
\begin{table}[!htb]
 
  \small
  
  \centering
  \begin{tabular}{|l|l|l|l|l|}
  \hline

    Models       & F-PPL&R-PPL&F-PPL&R-PPL \\
    			& SNLI&SNLI&COCO&COCO\\
    \hline
    Real  & 16.01 &\hspace{.2cm} -&67.05&\hspace{.2cm}-\\
    AAE  &74.56 &48.04 &83.34&257.21\\
    ARAE  &66.70 &315.58 &64.63&185.20\\
    IWGAN  &193.30 &53.96 &94.95&80.99\\
    \hline
	Soft-GAN & 110.95 & 142.13&61.42&170.17\\
	LATEXT-GAN I  & 86.24& 56.78  &\textbf{51.39}&158.21\\
    LATEXT-GAN II  & \textbf{53.22}& 144.80&63.32& 203.39 \\
    LATEXT-GAN III  &54.80& 143.42&71.60&214.76 \\
    \hline
  \end{tabular}
  \caption{Forward (F) and Reverse (R) perplexity (PPL) results for the SNLI and COCO datasets using synthetic sentences of maximum length 15 and 20 respectively.
  }
  \label{ppl-snli}
\end{table}
\vspace{-.3cm}
\subsection{Human Evaluation}
The subjective judgments of the synthetic sentences of the models trained using the COCO dataset with maximum sentence length of size 20 is reported in Table~\ref{human_eval}. We used 20 different random synthetic sentences generated by using the last iteration of each model and gave them to a group of 5 people. We asked them to rate the sentences based on a 5-point Likert scale according to their fluency. The raters are asked to score 1 which corresponds to gibberish, 3 corresponds to understandable but ungrammatical, and 5 correspond to naturally constructed and understandable sentences~\cite{Cifka2018}. From Table~\ref{human_eval}, we can note that the proposed LATEXT-GAN III approach get the higher rate compare to the other approaches. From all the above different evaluations,  we can note that the synthetic sentences by using the LATEXT-GAN II and III approaches are more balanced in diversity and fluency compare to the other approaches. 
We also depicted some examples of the synthetic sentences 
%in Table 7
%~\ref{sent-coco}
%and 8 %~\ref{sent-snli} 
 for the COCO and the SNLI datasets 
 %respectively
 %in appendix A.
in Table~\ref{sent-coco} and~\ref{sent-snli} respectively.
\begin{table}[!htb]
  \small
    \centering
  \begin{tabular}{|l|l|}
  \hline

    Model       & Fluency   \\
    \hline
    Real      &4.32  \\
    AAE      &1.72  \\
    ARAE     &2.60  \\
    IWGAN    &1.58 \\
    
    \hline
    Soft-GAN  &1.66 \\
    LATEXT-GAN I  &1.82\\
    LATEXT-GAN II  &2.48\\
    LATEXT-GAN III  &\textbf{2.70}\\
    
    \hline
  \end{tabular}
  \caption{Human Evaluation on the synthetic sentences}
  \label{human_eval}
\end{table} 
 %\clearpage
 %\appendix 

%\section{Example of synthetic text generated by different models}
%\numberwithin{table}{section}
%\label{sec:appendix}

% %%NEW TABLE
% \begin{table}[!htb]
%   \small
%     \centering
%   \begin{tabular}{|l|l|}
%   \hline

%     Model       & Fluency   \\
%     \hline
%     Real      &4.57  \\
%     AAE      & 1.71  \\
%     ARAE     & 2.64  \\
%     IWGAN    &1.95 \\
    
%     \hline
%     Soft-GAN  & 1.65 \\
%     LATEXT-GAN I  & 1.78\\
%     LATEXT-GAN II  & 2.35\\
%     LATEXT-GAN III  &\textbf{2.59 }\\
    
%     \hline
%   \end{tabular}
%   \caption{New Human Evaluation on the synthetic sentences}
%   \label{human_eval}
% \end{table}

\begin{table*}[!htb]

  \centering
  \scalebox{0.92}{
  \begin{tabular}{|p{5.5cm}|p{5.5cm}|p{5.5cm}|}
  
    \hline
          \hspace{1.75cm}ARAE       &       \hspace{1.75cm}{IWGAN}   &       \hspace{1.75cm}{Soft-GAN}   \\ 
    \hline
    a motorcycle parked outside of a counter street .&a small hang bathroom with a  park and side . &a man is sitting with a counter counter .\\ 
  a plane scene with focus on a toilet .         &a group of kitchen sits near a sliding .& a dog with a people above a street .         \\
 a cat standing down the of a bathroom . & a picture from a giraffe in a garlic .&  a blue of cat sitting in a red wall .                   \\
 a bathroom with a toilet, a wall and .   &a man shorts on a large table over .& a woman is looking with a red .     \\
 a white cat sits in the kitchen bowl .    & a car is hidden from a small kitchen .& a dog is sitting in a bathroom .\\
\hline
\hspace{1.75cm}LATEXT-GAN I       &       \hspace{1.5cm}{LATEXT-GAN II}   &       \hspace{1.5cm}{LATEXT-GAN III}   \\ 
    \hline
    a bathroom bathroom with with toilet and and mirror . &a bathroom with a sink and and white plane .& there are a park with ride on around display . \\
     a man kitchen next to a kitchen cabinets and .& a kitchen filled with wooden sink and a large window .& a kitchen with white cabinets and a black tub .\\ 
     a man standing at a kitchen in a . &a person is parked on a city road . &  a bathroom has a sink and a toilet . \\
     a man in in front a a bathroom with . &a cat sitting on a bench of a city street .&a woman riding a bike with a large on .           \\
    a plane is sitting with the sky at .&a group of people sitting on a city bench . & a picture of a kitchen with a a skies .  
 \\
    \hline
  \end{tabular}}
  
  %\onecolumn
  \caption{Example generated sentences with models trained using COCO dataset with maximum sentence length of size 20}
  \label{sent-coco}
\end{table*}

\begin{table*}[!htb]

  \centering
    \scalebox{0.9}{
  \begin{tabular}{|p{5.5cm}|p{5.5cm}|p{5.5cm}|}
    \hline
          \hspace{1.75cm}ARAE       &       \hspace{1.75cm}{IWGAN}   &       \hspace{1.75cm}{Soft-GAN}   \\ 
    \hline
    A little girl is playing outside .&A car is sitting in a house .        &A children of a blue a shirt while a .      
\\ 
  A woman is near the water .                 &Someone are Indian in his trinkets . & A little girl is looking while a a .       
\\
 A group of people are sitting .         & A woman is outside .          &  A couple of people are in white a a .      \\
 The lady is in the water .        &the man is in the house .         & A couple in a blue white is is standing .      
\\
 A lady sitting on a bench .            & The kids are laughing Bull .           & A man in a white shirt outside a a .     
\\
\hline
\hspace{1.75cm}LATEXT-GAN I       &       \hspace{1.5cm}{LATEXT-GAN II}   &       \hspace{1.5cm}{LATEXT-GAN III}   \\ 
    \hline
    A woman is holding a .&People are walking in the park .& A group of people are sitting in a field .\\
    A man is playing a .& The girl is playing with a a .& A boy is outside in a field .\\ 
    A man is standing on an beach .&There are motorcycles outside .& A man is riding a bike in the water .\\
 A man is walking in . &The young boy is wearing a red shirt .&A woman is riding a bike .    \\
    A little girl is playing in the .&A woman is walking on the park . & The man is standing .\\
    \hline
  \end{tabular}}
  \caption{Example generated sentences with models trained using SNLI dataset with maximum sentence length of size 15}
  \label{sent-snli}
\end{table*}

\vspace{-0.2cm}
\section{Conclusion and Future Work}
\label{conclusion}
In this paper, we introduced Soft-GAN as a new solution for the main bottleneck of using GAN for generating text, which is the discontinuity of text. 
This is based on applying soft-text to the GAN discriminator instead of the one-hot representation in the traditional approach. We also introduced three LATEXT-GAN approaches by combining the reconstructed output (soft-text) of an auto-encoder to the latent code-based GAN approaches (AAE and ARAE) for text generation. LATEXT-GAN I is formed on top of AAE method. LATEXT-GAN II and III approaches were formed based on ARAE. The LATEXT-GAN I and II approaches used separate discriminators for the synthetic text and the synthetic code discriminations. The LATEXT-GAN III used the combination of the soft-text and the latent code to compare with the concatenation of the synthetic text and the synthetic code by using a single discriminator. %We showed the power of combining the latent code and the soft-text over the well-known tricks to work around the discontinuity of text for GANs.
We evaluated the proposed approaches over the SNLI and the COCO datasets using subjective and objective evaluation metrics. 
The results of the experiments are consistent with different evaluation metrics. We showed the superiority of our proposed techniques, especially the LATEXT-GAN III method over other conventional GAN-based techniques which does not require pre-training.   
Finally, we summarize our plan for future work in the following: 
\begin{enumerate}
\item We trained the GAN using WGAN-GP approach. Spectral normalization technique~\cite{Miyato2018} can be applied to stabilize the GAN training which could generate more diverse text in our settings.
\item The proposed approaches are the pure GAN-based techniques for text generation and they are not very powerful in generating long sentences. RL or self-attention~\cite{Zhang2018} techniques can be used as a tool to accommodate this weakness.    
\item We used the hidden code from the last time step of the encoder. Attention mechanism can be applied for decoding. Furthermore, the powerful transformer~\cite{Vaswani2017} can be applied for the auto-encoder which could improve the performance of the proposed approaches. Pre-train the auto-encoder before adversarial training can also improve the performance.

\end{enumerate}

%\twocolumn

\section*{Acknowledgments}
We would like to thank Professor Qun Liu, Chief Scientist of Speech and Language Computing, Huawei Noah's Ark Lab  for his valuable comments on the paper.
%The acknowledgments should go immediately before the references.  Do
%not number the acknowledgments section. Do not include this section
%when submitting your paper for review. \\

%\noindent {\bf Preparing References:} \\
%Include your own bib file like this:
%\verb|\bibliographystyle{acl_natbib}|
%\verb|\bibliography{naaclhlt2019}| 

%where \verb|naaclhlt2019| corresponds to a naaclhlt2019.bib file.
\bibliography{naaclhlt2019}
\bibliographystyle{acl_natbib}

\end{document}